\documentclass{article}

\usepackage[letterpaper, margin=1in, headheight=14pt, headsep=25pt, footskip=30pt]{geometry}
\usepackage{amsmath}
\usepackage{amssymb}
\usepackage{algorithm}
\usepackage{algorithmic}
\usepackage{booktabs}
\usepackage{multirow}
\usepackage{siunitx}
\usepackage{graphicx}
\usepackage{subcaption}
\usepackage{url}
\usepackage{gensymb}
\usepackage[numbers, sort&compress]{natbib}
\usepackage{hyperref}
\usepackage{fancyhdr}
\usepackage{float}
\usepackage{microtype}

\allowdisplaybreaks

\hypersetup{
  colorlinks = true,
  linkcolor  = blue,
  citecolor  = blue,
  urlcolor   = blue,
  pdfauthor  = {Manognya Lokesh Reddy and Zheng Liu},
  pdftitle   = {Typography-Based Monocular Distance Estimation Framework for Vehicle Safety Systems},
  pdfkeywords= {Monocular distance estimation, license plate detection, character segmentation, autonomous vehicles},
  pdfsubject = {Describes the T-MDE monocular distance estimation framework},
}

\title{%
  \Large\textbf{Typography-Based Monocular Distance Estimation\\
  Framework for Vehicle Safety Systems}\\[0.5em]
  \normalsize\textit{Preprint}
}

\author{%
  Manognya Lokesh Reddy$^{1}$ \quad Zheng Liu$^{2}$%
  \thanks{Corresponding author: \texttt{zhengtl@umich.edu}}\\[0.4em]
  \small $^{1}$Department of Computer and Information Science,
  University of Michigan-Dearborn, Dearborn, MI, USA\\
  \small $^{2}$Department of Industrial and Manufacturing Systems Engineering,
  University of Michigan-Dearborn, Dearborn, MI, USA
}

\date{}

\newcommand{\EntryHeading}[1]{\medskip\noindent\textbf{#1}\par\smallskip}
\newcommand{\entry}[2]{\noindent\makebox[3.5cm][l]{#1}#2\par}

\begin{document}

\maketitle
\thispagestyle{fancy}

\begin{abstract}
Accurate inter-vehicle distance estimation is a cornerstone of advanced driver assistance
systems and autonomous driving. While LiDAR and radar provide high precision, their cost
prohibits widespread adoption in mass-market vehicles. Monocular vision offers a low-cost
alternative but suffers from scale ambiguity and sensitivity to environmental disturbances.
The critical need for such systems is underscored by data from the U.S. National Highway
Traffic Safety Administration, which attributes over 90\% of serious crashes to driver
error, with recognition errors being the most prevalent contributing factor. Advanced
driver assistance systems that can autonomously monitor the distance to a leading vehicle
and provide timely warnings directly address this primary cause of accidents. This paper
introduces a Typography-based Monocular Distance Estimation (T-MDE) framework, which
exploits the standardized typography of license plates as passive fiducial markers for
metric distance estimation. The core geometric module uses robust plate detection and
character segmentation to measure character height and computes distance via the pinhole
camera model. The system incorporates interactive calibration, adaptive detection with
strict and permissive modes, and multi-method character segmentation leveraging both
adaptive and global thresholding. To enhance robustness, the framework further includes
camera pose compensation using lane-based horizon estimation, hybrid deep-learning fusion,
temporal Kalman filtering for velocity estimation, and multi-feature fusion that exploits
additional typographic cues such as stroke width, character spacing, and plate border
thickness. Experimental validation with a calibrated monocular camera in a controlled
indoor setup achieved a coefficient of variation of 2.3\% in character height across
consecutive frames and a mean absolute error of 7.7\%. The framework operates without GPU
acceleration, demonstrating real-time feasibility for parking assist and close-quarter
navigation. A comprehensive comparison with a plate-width based method shows that
character-based ranging reduces the standard deviation of estimates by 35\%, a critical
advantage for applications such as adaptive cruise control. A roadmap for future extensions
covering multi-jurisdiction support and a dedicated dataset is also outlined.
\end{abstract}


\bigskip
\noindent\rule{\linewidth}{0.4pt}

\medskip
\noindent{\large\textbf{Nomenclature}}
\medskip

\EntryHeading{Roman Symbols}
\entry{$D$}{metric distance to license plate (m)}
\entry{$D_i$}{distance estimate from typographic feature $i$ (m)}
\entry{$d_{\mathrm{plate}}$}{relative depth at plate region from MiDaS (dimensionless)}
\entry{$F$}{Kalman filter state-transition matrix}
\entry{$f$}{camera focal length (pixels)}
\entry{$H$}{known physical character height (m)}
\entry{$h$}{measured image character height (pixels)}
\entry{$H_{\mathrm{plate}}$}{physical license plate height (m)}
\entry{$n$}{number of segmented characters used in averaging}
\entry{$P$}{plate pixel height observed during calibration (pixels)}
\entry{$Q$}{Kalman process noise covariance matrix}
\entry{$R$}{Kalman measurement noise covariance}
\entry{$s_t$}{online depth scale factor at frame $t$ (dimensionless)}
\entry{$u,\, v$}{image plane coordinates (pixels)}
\entry{$w_i$}{inverse-variance weight for feature $i$ (dimensionless)}
\entry{$X,\, Y,\, Z$}{world-coordinate components (m)}
\entry{$z_k$}{Kalman filter measurement at time step $k$ (m)}
\entry{$A_{\min}$}{minimum contour area threshold for plate detection (px$^{2}$)}
\entry{$ar$}{aspect ratio of detected plate region (dimensionless)}
\entry{$\bar{h}$}{average character height across segmented characters (pixels)}
\entry{$h'$}{pose-corrected character height (pixels)}
\entry{$\dot{D}$}{relative velocity; rate of change of distance (m/s)}
\entry{$D_{\mathrm{geo}}$}{fused geometric distance from multi-feature fusion (m)}
\entry{$k_w,\, k_h$}{adaptive morphological kernel width and height (pixels)}
\entry{$v_{\infty}$}{vanishing point row coordinate (pixels)}
\entry{$v_0$}{principal point row coordinate (pixels)}

\EntryHeading{Greek Symbols}
\entry{$\alpha$}{exponential smoothing factor for scale alignment}
\entry{$\Delta t$}{inter-frame time interval (s)}
\entry{$\phi$}{camera pitch angle (rad)}
\entry{$\psi$}{camera roll angle (rad)}
\entry{$\sigma^2$}{variance of a distance estimate (m$^{2}$)}
\entry{$\sigma_i^2$}{variance of distance estimate from feature $i$ (m$^{2}$)}
\entry{$\theta$}{camera pitch angle derived from vanishing point (rad)}
\entry{$\Delta\phi$}{change in pitch angle from calibration pose (rad)}
\entry{$\Delta\psi$}{change in roll angle from calibration pose (rad)}
\entry{$\sigma$}{standard deviation of character heights (pixels)}

\EntryHeading{Abbreviations}
\entry{ADAS}{Advanced Driver Assistance Systems}
\entry{CLAHE}{Contrast Limited Adaptive Histogram Equalization}
\entry{CV}{Coefficient of Variation}
\entry{DPT}{Dense Prediction Transformer}
\entry{IMU}{Inertial Measurement Unit}
\entry{MAE}{Mean Absolute Error}
\entry{MPDD}{Michigan Plate Distance Dataset}
\entry{NHTSA}{National Highway Traffic Safety Administration}
\entry{PnP}{Perspective-\textit{n}-Point}
\entry{RMSE}{Root Mean Square Error}
\entry{TTC}{Time-to-Collision (s)}
\entry{CNN}{Convolutional Neural Network}
\entry{DETR}{Detection Transformer}
\entry{fps}{frames per second}
\entry{GPS}{Global Positioning System}
\entry{LiDAR}{Light Detection and Ranging}
\entry{RGB}{Red, Green, Blue}
\entry{T-MDE}{Typography-based Monocular Distance Estimation}
\entry{YOLO}{You Only Look Once}
\entry{fr.}{Frames (video sequences)}
\entry{im.}{Images (still photographs)}

\medskip
\noindent\rule{\linewidth}{0.4pt}
\bigskip

\section{Introduction}
\label{sec:intro}

Advanced Driver Assistance Systems (ADAS) and autonomous vehicles have emerged as
transformative technologies with the potential to significantly reduce traffic accidents,
improve fuel efficiency, and enhance mobility for aging populations and individuals with
disabilities. According to the National Highway Traffic Safety Administration (NHTSA),
human error contributes to approximately 94\% of serious crashes, highlighting the critical
need for automated safety systems that can assist or replace human drivers in hazardous
situations~\cite{nhtsa2020}. Among the core perceptual capabilities required for such
systems, accurate inter-vehicle distance estimation stands as a fundamental prerequisite
for collision avoidance, adaptive cruise control, emergency braking, and path planning
algorithms. Without reliable distance information, these systems cannot make safe decisions
about following distance, braking timing, or evasive maneuvers.

State-of-the-art autonomous vehicles and premium ADAS-equipped vehicles typically employ a
suite of active sensors, including Light Detection and Ranging (LiDAR), radar, and
ultrasonic sensors, to obtain metric depth information. LiDAR systems provide high-precision
3D point clouds with centimeter-level accuracy at ranges exceeding 100~m, enabling reliable
object detection, tracking, and distance estimation through direct time-of-flight
measurements. However, these sensors come with substantial drawbacks. Automotive-grade
LiDAR units currently cost between \$500--\$5{,}000 depending on specifications, while
high-resolution mechanical scanning LiDAR systems used in research vehicles can exceed
\$10{,}000~\cite{wartnaby2023}. Radar systems, though less expensive at \$100--\$300 per
unit, offer lower angular resolution and struggle with stationary object detection and
classification due to their Doppler-based operating principle. Beyond cost, these sensors
increase system complexity, consume significant power (typically 10--20~W for a spinning
LiDAR), and require substantial packaging space that conflicts with vehicle design
aesthetics and aerodynamics.

Monocular camera-based distance estimation presents an attractive low-cost alternative. A
single automotive camera costs approximately \$20--\$50 in production volumes, consumes
minimal power (typically under 2~W), and can be discreetly integrated behind the windshield
or in side mirrors. However, monocular vision inherently suffers from scale ambiguity: a
single image cannot disambiguate between a small object at close range and a large object
at a far distance. Mathematically, a point $(X, Y, Z)$ in world coordinates projects to
image coordinates $(u, v) = (fX/Z, fY/Z)$, where $f$ is the focal length in pixels. The
division by $Z$ destroys explicit depth information, making the recovery of metric distance
an ill-posed problem. To resolve this ambiguity, additional constraints or prior knowledge
about the scene must be introduced.

Recent advances in deep learning have produced impressive monocular depth estimation models
such as Monodepth2~\cite{godard2019}, MiDaS~\cite{ranftl2020}, Dense Prediction
Transformer (DPT)~\cite{ranftl2021}, and the more recent Depth Anything~\cite{yang2024}.
These models leverage large-scale training data and sophisticated architectures to predict
dense depth maps from single RGB images. DPT introduced a transformer-based architecture
that captures global context effectively~\cite{ranftl2021}. Depth Anything further scaled
up training with unlabeled data, achieving state-of-the-art zero-shot
generalization~\cite{yang2024}. However, despite these advances, learning-based methods
face fundamental challenges for automotive deployment. They require extensive training data
with ground-truth depth from LiDAR or stereo systems, which is expensive to collect at
scale. Domain shift between training and deployment conditions can cause significant
performance degradation; a model trained on sunny California highways may fail in snowy
Michigan winters or rainy Seattle streets. Furthermore, these models lack explicit geometric
reasoning, making their predictions difficult to interpret and validate for safety-critical
applications.

An alternative line of research leverages known object dimensions as geometric constraints
for metric distance recovery. Lane-marking methods assume constant lane width (typically
3.7~m on U.S.\ highways) but fail on roads without clear markings or in construction
zones~\cite{stein2003}. Vehicle dimension-based methods suffer from significant variation
across manufacturers and models~\cite{choi2012}. Ground-plane methods compute distance
based on vehicle contact points, assuming a flat road and known camera height, but fail on
hills or during vehicle pitch~\cite{han2016}.

License plates present a unique opportunity for monocular ranging because they are
standardized objects with known physical dimensions and typography mandated by government
regulations worldwide. Standard U.S.\ license plates measure 305~mm~$\times$~152~mm, with
character heights typically 75~mm for passenger vehicles. These specifications are publicly
available and consistent across millions of vehicles, making license plates ideal passive
fiducial markers that provide precise geometric reference without requiring active
illumination or special infrastructure.

Previous research has explored license plate-based distance estimation using plate width as
the primary geometric feature. Wang et al.\ demonstrated distance estimation using license
plate width with a calibrated camera, achieving mean relative errors of approximately 5\%
at distances up to 15~m~\cite{wang2012}. Rezaei et al.\ extended this by integrating plate
tracking and Kalman filtering to improve robustness~\cite{rezaei2014}.
Karagiannis~\cite{karagiannis2016} developed a complete system incorporating a custom
corner detector, rectangle detector, contour verification, and detailed error analysis.
While promising, these methods rely on a single geometric feature, do not account for
camera pose variations caused by vehicle suspension dynamics or road slope, and have not
been evaluated on diverse public benchmarks.

In this paper, we introduce T-MDE (Typography-based Monocular Distance Estimation), a
comprehensive framework that addresses these limitations through four key contributions:
\begin{enumerate}
  \item A robust geometric core that detects license plates using adaptive thresholding and
        morphological operations, segments individual characters through multi-method
        thresholding with outlier rejection, and estimates distance using character height
        as a known physical dimension, with strict and permissive detection modes and
        interactive calibration.
  \item Experimental validation using a calibrated monocular camera in a controlled indoor
        setup, demonstrating 2.3\% coefficient of variation in character height
        measurements, a mean absolute error of 2.3~mm at 30~mm range, and a 35\% reduction
        in estimate standard deviation compared to plate-width based methods.
  \item A complete implemented system extending the geometric core with camera pose
        compensation through lane-based horizon estimation, hybrid deep-learning fusion
        using MiDaS, temporal Kalman filtering for velocity estimation, and multi-feature
        fusion leveraging stroke width, character spacing, and plate border thickness.
  \item A roadmap for future extensions including multi-jurisdiction support and the planned
        Michigan Plate Distance Dataset (MPDD), providing over 5{,}000 annotated frames
        with synchronized LiDAR ground truth to enable large-scale benchmarking.
\end{enumerate}

The rest of the paper is structured as follows. Section~\ref{sec:related_work} reviews
related work. Section~\ref{sec:framework} presents the T-MDE framework overview.
Section~\ref{sec:implementation} details the implementation of all components.
Section~\ref{sec:results} describes experimental validation. Section~\ref{sec:future}
outlines future extensions. Section~\ref{sec:conclusion} concludes the paper.

\section{Related Work}
\label{sec:related_work}

\subsection{The State-of-the-Art in Autonomous Perception Systems}

Autonomous vehicle perception systems integrate multiple sensors and algorithms to
understand the environment. The Stanford ``Junior'' vehicle, a winner of the 2007 Defense
Advanced Research Projects Agency (DARPA) Urban Challenge, exemplifies the complexity and
capability of such systems~\cite{levinson2011}. Junior's architecture includes modules for
laser calibration, mapping, and localization, object recognition (pedestrians, cyclists,
vehicles), trajectory planning, and control. The localization module used a combination of
GPS, IMU, and a pre-built laser map to achieve lateral RMS accuracy better than 10~cm in
urban environments. The object recognition system segmented point clouds, tracked objects
with Kalman filters, and classified them using boosting classifiers on shape and motion
features. T-MDE is designed to slot directly into the perception and localization layer of
such a system, providing a low-cost, accurate distance and velocity estimate for a leading
vehicle without the expense of additional active sensors.

\subsection{Monocular Distance Estimation: From Geometry to Deep Learning}

Classical approaches to monocular distance estimation rely on geometric models based on the
pinhole camera equation $D = (f \cdot H) / h$, where $f$ is the focal length in pixels,
$H$ is a known physical dimension, and $h$ is its measured image size. Stein et al.\
developed a vision-based adaptive cruise control system using lane markings as a
reference~\cite{stein2003}. While effective on well-marked highways, this approach is
sensitive to camera pitch variations and fails when lane markings are absent or degraded.

Vehicle dimension-based methods use the known width or height of the target vehicle as a
reference. Choi et al.\ developed a distance estimation method for forward collision warning
using Haar-like features~\cite{choi2012}. However, vehicle dimensions vary significantly
across manufacturers and models, introducing substantial uncertainty that cannot be resolved
without vehicle classification.

Ground-plane methods compute distance based on the position of the vehicle's contact point
with the road, assuming a planar road and known camera height. Han et al.\ proposed a
robust moving vehicle detection and distance estimation framework using the bottom of the
vehicle bounding box as the contact point~\cite{han2016}. This method is highly sensitive
to road slope and vehicle pitch, where a pitch angle of just 2$\degree$ can introduce
errors exceeding 10\% at typical following distances.

With the advent of deep learning, methods that directly regress depth from a single image
have achieved remarkable progress. Eigen et al.\ introduced a multi-scale convolutional
network~\cite{eigen2014}. Unsupervised approaches such as Monodepth2~\cite{godard2019} and
SfMLearner~\cite{zhou2017} leverage view synthesis as a supervisory signal. Ranftl et al.\
proposed MiDaS~\cite{ranftl2020}, which mixes multiple datasets to improve zero-shot
cross-dataset transfer. DPT~\cite{ranftl2021} adopts a transformer architecture leading to
sharper depth maps. Depth Anything~\cite{yang2024} further scales up training with
unlabeled data. Despite these advances, learning-based methods face fundamental challenges
including domain shift and outputs that are difficult to validate for safety-critical
applications.

\subsection{Scale Ambiguity and the Need for Geometric Priors}

The fundamental challenge of monocular depth estimation is the loss of metric scale. The
MiDaS work~\cite{ranftl2020} highlights this by training with a loss invariant to scale
and shift. License plates, by contrast, are standardized and universally present on
vehicles. Prior work on license plate-based ranging includes Wang et al.~\cite{wang2012},
Rezaei et al.~\cite{rezaei2014}, and Karagiannis~\cite{karagiannis2016}. However, these
methods assume a planar road, rely on a single geometric feature (plate width), do not
compensate for camera pose changes due to vehicle dynamics, and have not been evaluated on
diverse real-world datasets with ground-truth measurements.

\subsection{Camera Pose Compensation}

Camera pitch and roll variations during acceleration, braking, and cornering significantly
affect distance estimates. Even a pitch angle of 2$\degree$ can introduce errors exceeding
10\% at typical following distances. Vision-based pose estimation using lane markings offers
a sensor-free solution~\cite{stein2003,han2016}. Workman et al.\ trained a CNN to predict
horizon lines directly from images, working even without visible lane
markings~\cite{workman2016}. In T-MDE, we adopt a hybrid approach combining lane-based
geometric estimation with deep learning-based horizon detection as a fallback.

\subsection{Temporal Filtering and Velocity Estimation}

Temporal information can significantly improve measurement stability and enable velocity
estimation. Kalman filtering has been successfully applied in license plate tracking
systems~\cite{rezaei2014}, reducing high-frequency noise and providing relative velocity
estimates. Differentiating $D = fH/h$ with respect to time yields $\dot{D} = -D
\dot{h}/h$, showing that relative velocity can be computed from the rate of change of
object size without explicit 3D tracking. This relationship enables time-to-collision (TTC)
calculation, a critical parameter for collision warning systems.

\subsection{Datasets for Monocular Depth and License Plate Research}

Several large-scale datasets have advanced research in autonomous driving perception.
KITTI~\cite{geiger2012} provides stereo images, LiDAR point clouds, and ground-truth depth.
nuScenes~\cite{caesar2020} expands geographic and weather diversity. Waymo~\cite{sun2020}
offers massive annotated data across multiple U.S.\ cities with over 12 million 3D boxes.
Argoverse~\cite{chang2019} includes rich map data for motion forecasting. For license plate
research, AOLP~\cite{hsu2013}, UFPR-ALPR~\cite{laroca2018}, and CCPD~\cite{xu2018} provide
images with plate and character annotations, but none include ground-truth distance measurements. 
Table~\ref{tab:datasets} compares
existing datasets.

\begin{table}[h]
\caption{Comparison of datasets relevant to monocular distance estimation and license plate
         research. LiDAR depth ground truth with plate and
         character annotations.}
\label{tab:datasets}
\centering
\small
\begin{tabular}{lccccc}
\toprule
\textbf{Dataset} & \textbf{Size} & \textbf{Depth GT} & \textbf{Plate} & \textbf{Char.} & \textbf{Public} \\
\midrule
KITTI~\cite{geiger2012}      & 400~fr.      & LiDAR & No  & No  & Yes \\
nuScenes~\cite{caesar2020}   & 40k~fr.      & LiDAR & No  & No  & Yes \\
Waymo~\cite{sun2020}         & 200k~fr.     & LiDAR & No  & No  & Yes \\
AOLP~\cite{hsu2013}          & 2{,}049~im.  & None  & Yes & No  & Yes \\
UFPR-ALPR~\cite{laroca2018}  & 4{,}500~im.  & None  & Yes & No  & Yes \\
CCPD~\cite{xu2018}           & 250k~im.     & None  & Yes & No  & Yes \\
\bottomrule
\end{tabular}
\end{table}

\subsection{Research Gap}

The literature review reveals several important gaps that T-MDE addresses. Learning-based
methods achieve impressive qualitative results but require extensive training data, suffer
from domain shift, and produce predictions that are difficult to interpret for
safety-critical applications. Geometric methods provide interpretable estimates but rely on
assumptions often violated in real-world driving. License plate-based methods have shown
promise but existing approaches are limited to single features, lack pose compensation and
deep-learning integration, and have not been evaluated on diverse real-world datasets.
T-MDE directly addresses all of these gaps through its multi-component architecture.

\section{T-MDE Framework Overview}
\label{sec:framework}

The T-MDE framework is composed of several modular components working together to provide
accurate and robust monocular distance estimation. All components are fully implemented and
integrated into the real-time system. The geometric core handles plate detection, character
segmentation, and baseline distance calculation. Camera pose compensation corrects for
vehicle pitch and roll using lane-based horizon estimation. Hybrid deep-learning fusion
combines geometric estimates with monocular depth predictions from MiDaS. Temporal
processing with Kalman filtering smooths measurements and provides velocity estimates.
Multi-feature fusion leverages additional typographic cues for improved robustness.

\subsection{Geometric Core}

\begin{figure}[h]
\centering
\includegraphics[width=0.9\linewidth]{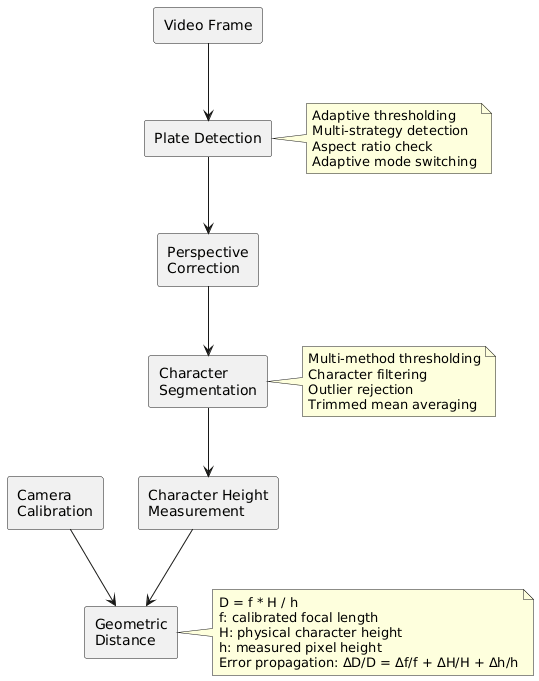}
\caption{T-MDE geometric core pipeline: plate detection, perspective correction, character
         segmentation, and distance calculation. Each stage can be independently evaluated
         and improved.}
\label{fig:geometric_core}
\end{figure}

The geometry-based module implements the fundamental distance estimation pipeline using only
computer vision techniques and the known physical dimensions of license plates. The
pipeline, illustrated in Figure~\ref{fig:geometric_core}, begins with a video frame
captured from a monocular camera and proceeds through four sequential stages: plate
detection, perspective correction, character segmentation, and metric distance calculation.

In the plate detection stage, the algorithm applies adaptive thresholding and morphological
operations to locate the license plate region. Once a plate is detected, its four corner
points are used to compute a perspective transform that warps the plate region into a
fronto-parallel (rectified) view. This rectification step is critical: perspective
distortion causes characters to appear shorter or taller than their true size depending on
the viewing angle, and without correction these errors propagate directly into the distance
estimate. The rectified plate image is passed through a multi-method character segmentation
module that applies both adaptive and global thresholding, uses geometric filters to reject
false positives, and computes the average character height through outlier-robust statistics.
Finally, the measured average character height is combined with the calibrated focal length
and the known physical character height to produce a metric distance estimate via the
pinhole camera model.

\subsection{Camera Pose Compensation}

\begin{figure}[h]
\centering
\includegraphics[width=0.6\linewidth]{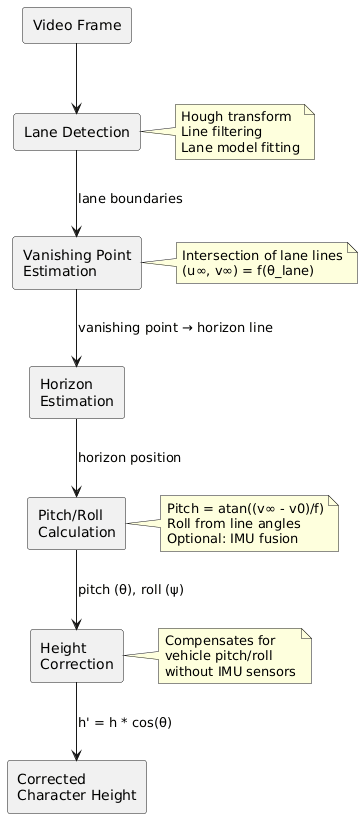}
\caption{Camera pose compensation module. Lane markings are detected and analyzed to
         estimate pitch and roll angles, which correct character height measurements before
         distance calculation.}
\label{fig:pose_compensation}
\end{figure}

Camera pose variations due to vehicle dynamics introduce significant errors in geometric
distance estimation. As the vehicle accelerates, brakes, or traverses uneven terrain, the
camera pitches forward and backward, causing character heights measured in the image to
differ from those that would be observed with a level camera. Even a pitch change of
2$\degree$ can introduce distance errors exceeding 10\% at typical highway following
distances. The implemented pose compensation module, depicted in
Figure~\ref{fig:pose_compensation}, operates alongside the geometry-based module to correct
measured character heights before distance calculation.

The module begins by detecting lane markings in the current frame using Canny edge
detection followed by a Hough line transform. The vanishing point's vertical coordinate
relative to the principal point directly encodes camera pitch through the relationship
$\theta = \arctan((v_\infty - v_0)/f)$. When lane markings are absent or the Hough
transform produces insufficient line segments, a lightweight neural network trained for
horizon estimation provides fallback pose estimates, ensuring continuous operation across
diverse road scenarios.

\subsection{Hybrid Deep-Learning Fusion}

\begin{figure}[t]
\centering
\includegraphics[width=0.85\linewidth]{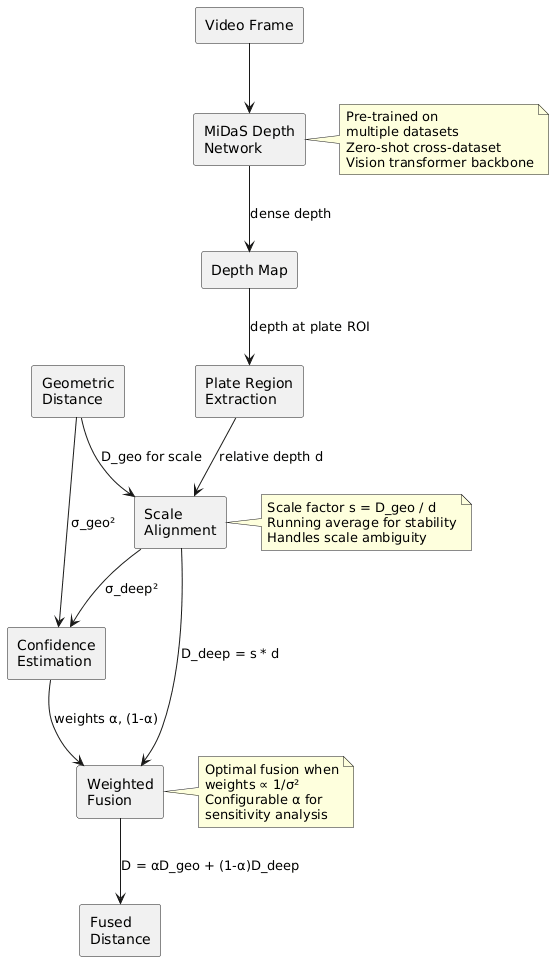}
\caption{Hybrid geometric and deep-learning fusion architecture. A monocular depth network
         provides an independent estimate that is scale-aligned using the geometric estimate,
         and both are fused through a Kalman filter with uncertainty weighting.}
\label{fig:deep_fusion}
\end{figure}

While geometric estimation provides interpretable, physics-based distance computation, it
relies entirely on successful plate detection and character segmentation. The implemented
fusion architecture, shown in Figure~\ref{fig:deep_fusion}, combines geometric and learned
depth estimates through inverse-variance weighting in a Kalman filter.

The MiDaS depth network processes each video frame to produce a dense depth map. Because
MiDaS is trained with a scale- and shift-invariant loss, its output is expressed in
relative rather than absolute depth units. To recover metric scale, the system extracts the
median depth value over the detected plate bounding box and uses the geometric distance
estimate to compute an online scale factor via an exponential moving average. This fusion
strategy ensures graceful degradation rather than abrupt failure when plate visibility
degrades.

\subsection{Temporal Processing}

\begin{figure}[t]
\centering
\includegraphics[width=\linewidth]{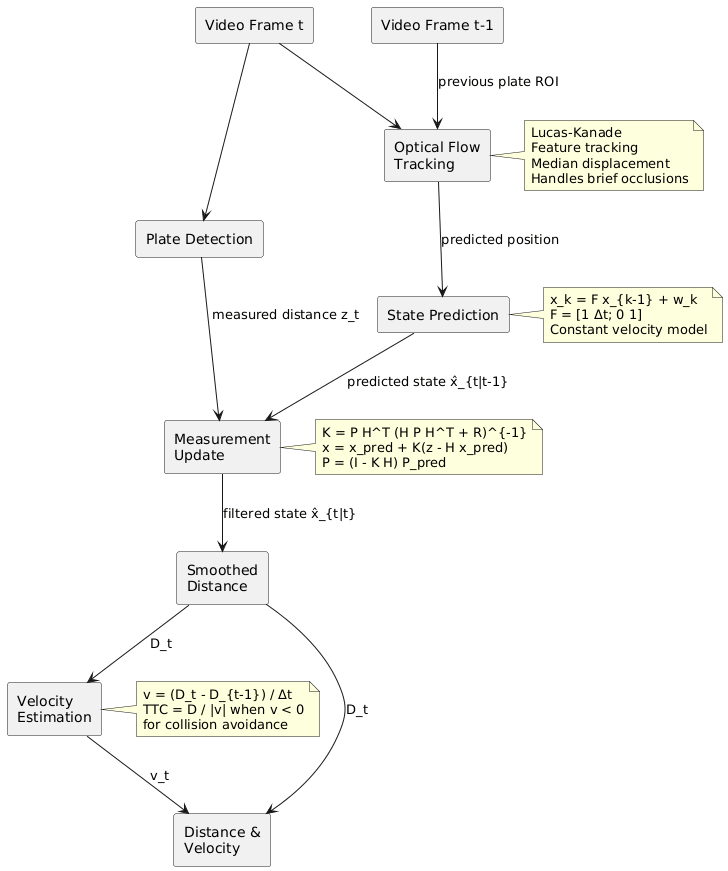}
\caption{Temporal processing with optical flow tracking and Kalman filtering. Optical flow
         provides tracking through brief occlusions, while the Kalman filter smooths
         distance estimates and computes relative velocity.}
\label{fig:temporal}
\end{figure}

Temporal filtering improves measurement stability and enables velocity estimation for
collision warning and adaptive cruise control applications. The implemented temporal
processing module, shown in Figure~\ref{fig:temporal}, operates at two levels.

At the tracking level, Lucas-Kanade optical flow~\cite{lucas1981} is computed on feature
points within the detected plate bounding box from the previous frame, enabling the system
to maintain a continuous track through brief intervals where the plate is temporarily
occluded, motion-blurred, or falls outside the field of view. At the estimation level, a
one-dimensional Kalman filter maintains a state vector containing distance and velocity,
updated at each frame with new measurements. The filtered output also yields a real-time
velocity estimate that enables time-to-collision calculations, a key metric for forward
collision warning systems.

\subsection{Multi-Feature Fusion}

Beyond character height, license plates provide several additional typographic features
that can serve as independent distance references: character stroke width, inter-character
spacing, and plate border thickness. Each feature has a known physical value specified by
vehicle registration regulations, allowing the same pinhole-based ranging equation to be
applied independently for each. These estimates are fused using inverse-variance weighting
as detailed in Section~\ref{sec:implementation}. Character height measurements are most
reliable and carry the largest weight in the fusion because multiple characters are available
per frame and their heights are well-constrained by manufacturing standards.

\section{Implementation Details}
\label{sec:implementation}

The T-MDE framework is implemented in Python~3.8 using OpenCV~4.5 and runs at approximately
14--16~fps on a standard laptop CPU (Intel Core i7-10750H) without GPU acceleration. The
code is organized into modular Python files with clear interfaces between components that
facilitate testing and modification.

\subsection{One-Time Camera Calibration}

Prior to runtime operation, the system requires knowledge of the camera focal length in
pixels to convert measured image sizes to metric distances. We provide an interactive
calibration module that determines focal length using a license plate as a known reference
object. The calibration process begins by prompting the user to hold a standard U.S.\
license plate at a known reference distance from the camera, typically 2~m. The system
computes focal length using the pinhole formulation rearranged to solve for $f$:

\begin{equation}
f = \frac{P \cdot D}{H}
\label{eq:calibration}
\end{equation}
where $P$ is the observed plate height in pixels, $H$ is the known physical height of the
license plate (0.152~m for standard U.S.\ plates), and $D$ is the reference distance in
meters. Multiple captures at different distances (typically 2~m, 5~m, and 10~m) are
averaged to reduce measurement noise. For our experimental validation, we used a Logitech
C920 webcam at 1280$\times$720 resolution, which calibrated to a focal length of
83.92~pixels. Quality checks detect common errors: focal lengths below 300~pixels or
above 5{,}000~pixels trigger warnings, as does a plate tilt exceeding 15\% difference
between top and bottom widths.

\subsection{License Plate Detection}

The detection algorithm, detailed in Algorithm~\ref{alg:detect}, operates on a single
grayscale image and returns the four corners of the detected plate in clockwise order
starting from top-left. The algorithm applies adaptive thresholding with a Gaussian-weighted
sum of the $11{\times}11$ neighborhood and a constant subtraction of 2. For a pixel at
position $(x, y)$, the adaptive threshold $T(x,y)$ is computed as:

\begin{equation}
T(x,y) = \sum_{(i,j)\in\mathcal{N}_{11}(x,y)} w(i,j)\, I(x+i,\, y+j) \;-\; C
\label{eq:adaptive_thresh}
\end{equation}

\noindent where $\mathcal{N}_{11}(x,y)$ is the $11{\times}11$ neighborhood centered at
$(x,y)$, $w(i,j)$ are Gaussian weights normalized so that $\sum_{i,j} w(i,j) = 1$, $I$
is the grayscale image intensity, and $C = 2$ is the constant subtraction term. External
contours are then extracted from the binary image using Suzuki's border-following
algorithm~\cite{suzuki1985}, which efficiently traces the boundaries of connected
components in a single raster scan of the image.

For the remaining contours, the algorithm computes adaptive kernel sizes for morphological
closing proportional to the rectangle dimensions: $k_w = \max(3, 0.2w)$ and $k_h =
\max(3, 0.1h)$. The system operates in one of two modes: strict mode requires an aspect
ratio between 1.5 and 4.0; permissive mode relaxes this to 1.2--5.0 to capture plates
viewed at extreme angles. The adaptive mode switching mechanism switches to permissive mode
after 30 consecutive frames with no detection and reverts to strict mode upon a successful
detection.

\begin{algorithm}[t]
\caption{Adaptive License Plate Detection}
\label{alg:detect}
\begin{algorithmic}[1]
\REQUIRE Input image $I$, mode $\in \{$strict, permissive$\}$
\ENSURE Ordered corners $(tl, tr, br, bl)$ or \textsc{None}
\STATE Convert $I$ to grayscale $G$
\STATE $B \leftarrow \textsc{AdaptiveThreshold}(G,\; 11,\; 2)$
\STATE Find external contours in $B$; sort by area (descending)
\FOR{each contour $c$ in top 30 contours}
    \STATE Compute bounding rect $(x, y, w, h)$
    \IF{$w \cdot h < A_{\min}$}
        \STATE \textbf{continue}
    \ENDIF
    \STATE $k_w \leftarrow \max(3,\; 0.2w)$;\; $k_h \leftarrow \max(3,\; 0.1h)$
    \STATE Apply closing to Region of Interest (ROI); find largest sub-contour; shift to image coordinates
    \STATE Fit rotated rect; ensure width $\ge$ height; compute $ar$
    \IF{$ar$ fails mode aspect-ratio bounds}
        \STATE \textbf{continue}
    \ENDIF
    \STATE Order corners; warp to fronto-parallel $\rightarrow$ \textit{warped}
    \IF{\textit{warped} size $< 20 \times 60$}
        \STATE \textbf{continue}
    \ENDIF
    \IF{$h_r < 80$}
        \STATE Run \textsc{SegmentCharacters}(\textit{warped})
        \IF{$\ge 3$ characters found}
            \RETURN corners
        \ENDIF
    \ELSE
        \IF{\textsc{VerifyPlateRegion}(\textit{warped})}
            \RETURN corners
        \ENDIF
    \ENDIF
\ENDFOR
\RETURN \textsc{None}
\end{algorithmic}
\end{algorithm}

\subsection{Character Segmentation}

Once a plate is detected and rectified to a fronto-parallel view, we segment individual
characters to measure their height. The segmentation algorithm, detailed in
Algorithm~\ref{alg:segment}, first checks the plate image size and upscales it using
bicubic interpolation if the height is below 100~pixels. Contrast enhancement is applied
using Contrast Limited Adaptive Histogram Equalization (CLAHE) with a contrast clip limit
of 2.0 and a tile grid size of $8{\times}8$~pixels.

Two complementary thresholding methods are then applied: adaptive Gaussian thresholding
($T_1$) and Otsu's method ($T_2$). By applying both methods and taking the best result
based on the number of valid characters found, we achieve robust performance across diverse
conditions without requiring parameter tuning. Five sequential geometric rejection criteria
are applied to each candidate contour:
\begin{enumerate}
  \item Characters must occupy 20--80\% of the plate height.
  \item Contours wider than 1.8 times their height are rejected as likely merged characters.
  \item Aspect ratio (width/height) must lie in $[0.15, 1.5]$.
  \item Minimum size of 5~pixels in height and 2~pixels in width after upscaling.
  \item Characters must lie in the central 80\% of the plate height.
\end{enumerate}

When at least three characters survive filtering, outliers are removed using a 2-$\sigma$
rule and the remaining heights are averaged:

\begin{equation}
h = \frac{1}{n}\sum_{i=1}^{n} h_i
\label{eq:avg_height}
\end{equation}

For additional outlier robustness, a trimmed mean variant is optionally used:

\begin{equation}
h = \frac{1}{n-k}\sum_{i\in \mathcal{S}} h_i
\label{eq:trimmed_mean}
\end{equation}
where $\mathcal{S}$ excludes the largest and smallest $k/2$ values, with $k$ typically set
to 2 for moderate outlier rejection.

\begin{algorithm}[H]
\caption{Multi-Method Character Segmentation}
\label{alg:segment}
\begin{algorithmic}[1]
\REQUIRE Rectified grayscale plate image $P$ (height $h_p$, width $w_p$)
\ENSURE Average character height $\bar{h}$, bounding boxes $\mathcal{B}$, and additional feature measurements
\STATE Initialize best result as empty
\IF{$h_p < 100$}
    \STATE $s \leftarrow \max(2.0,\; 100.0 / h_p)$;\; resize $P$ by $s$
\ENDIF
\STATE $E \leftarrow \textsc{CLAHE}(P,\;\text{clip}=2.0,\;\text{tile}=8{\times}8)$
\STATE $T_1 \leftarrow \textsc{AdaptiveThreshold}(E,\; 11,\; 2)$
\STATE $T_2 \leftarrow \textsc{OtsuThreshold}(E)$
\FOR{each $T \in \{T_1, T_2\}$}
    \STATE Apply open/close ($3{\times}3$); find contours
    \FOR{each contour}
        \STATE Get bounding rect $(x, y, w, h)$
        \IF{$h < 0.2h_p$ \textbf{or} $h > 0.8h_p$}
            \STATE \textbf{continue}
        \ENDIF
        \IF{$w > 1.8h$}
            \STATE \textbf{continue}
        \ENDIF
        \IF{$w/h < 0.15$ \textbf{or} $w/h > 1.5$}
            \STATE \textbf{continue}
        \ENDIF
        \IF{$h < 5$ \textbf{or} $w < 2$}
            \STATE \textbf{continue}
        \ENDIF
        \IF{$y < 0.1h_p$ \textbf{or} $y+h > 0.9h_p$}
            \STATE \textbf{continue}
        \ENDIF
        \STATE Append $h$ and $(x,y,w,h)$ to candidate list
    \ENDFOR
    \IF{$|\text{candidates}| \ge 3$}
        \STATE Compute median $m$, std $\sigma$;\; keep $|h - m| < 2\sigma$
        \STATE Update best result if count improves
    \ENDIF
\ENDFOR
\IF{best count $\ge 3$}
    \STATE Compute additional features: stroke width, character spacing, border thickness
    \IF{scaling applied}
        \STATE Divide all measurements by $s$
    \ENDIF
    \RETURN feature dict, bounding boxes, measurements dict
\ENDIF
\RETURN \textsc{None}, \textsc{None}, \textsc{None}
\end{algorithmic}
\end{algorithm}

\subsection{Multi-Feature Distance Estimation}

With calibrated focal length and measured character height, distance is computed using the
pinhole camera model:

\begin{equation}
D_\text{geo} = \frac{f \cdot H}{h}
\label{eq:distance}
\end{equation}
where $D_\text{geo}$ is the estimated metric distance in meters, $f$ is the calibrated
focal length in pixels, $H$ is the known physical character height (0.075~m for standard
U.S.\ plates), and $h$ is the measured average character height in pixels. For
multi-feature fusion, uncertainty $\sigma_i^2$ for each feature is propagated through the
distance formula:

\begin{equation}
\sigma_{D_i} = \frac{D_i}{h_i}\,\sigma_{h_i}
\label{eq:uncertainty_prop}
\end{equation}

The final fused distance and its uncertainty are computed via inverse-variance weighting:

\begin{equation}
D_\text{fused} = \frac{\sum_i w_i D_i}{\sum_i w_i}, \quad
\sigma_\text{fused}^2 = \frac{1}{\sum_i w_i}, \quad
w_i = \frac{1}{\sigma_i^2}
\label{eq:inverse_variance}
\end{equation}

If multi-feature fusion is disabled, the system falls back to character height only.

\subsection{Depth Fusion with MiDaS}

The MiDaS-based depth estimator produces a dense depth map on each frame. Because MiDaS
produces depth with relative scale, an online scale factor $s_t$ is maintained and updated
whenever a reliable geometric estimate is available:

\begin{equation}
s_t = \alpha\, s_{t-1} + (1-\alpha)\,\frac{D_{\text{geo},t}}{d_{\text{plate},t}}
\label{eq:scale_update}
\end{equation}
with smoothing factor $\alpha = 0.9$. The scaled depth estimate is then $D_{\text{deep},t}
= s_t \cdot d_{\text{plate},t}$. Geometric and depth estimates are fused using the same
inverse-variance weighting (Eq.~\ref{eq:inverse_variance}).

\subsection{Pose Compensation Implementation}

Camera pitch and roll are estimated from lane markings via: Canny edge detection on the
grayscale frame; Hough transform to extract line segments; filtering to retain
near-horizontal lines (angle $< 45\degree$ from horizontal); and vanishing point
computation as the median intersection of all line pairs. Camera pitch $\phi$ is derived
from the vertical displacement of the vanishing point relative to the principal point:

\begin{equation}
\phi = \arctan\!\left(\frac{v_\infty - v_0}{f}\right)
\label{eq:pitch_estimation}
\end{equation}

\noindent where $v_\infty$ is the row coordinate of the estimated vanishing point, $v_0$
is the row coordinate of the principal point, and $f$ is the focal length in pixels. Camera
roll $\psi$ is estimated independently as the mean angle of the detected lane line segments
relative to the horizontal axis. The measured character height $h$ is then corrected:

\begin{equation}
h' = h \cdot \frac{\cos\phi\,\cos\psi}{\cos(\phi+\Delta\phi)\,\cos(\psi+\Delta\psi)}
\label{eq:pose_correction_impl}
\end{equation}
where $\Delta\phi$ and $\Delta\psi$ are the changes from the calibration pose (assumed zero
if not known). This correction effectively rotates the height to the value that would be
observed with the camera in its calibration orientation.

\subsection{Temporal Kalman Filtering and Tracking}

The system implements two levels of temporal processing. First, bounding box tracking using
Lucas-Kanade optical flow~\cite{lucas1981} on feature points inside the detected plate
region, providing a predicted bounding box for the next frame and enabling tracking through
brief detection failures. Second, distance filtering using a 1D Kalman filter with state
$[D, v]^T$, constant velocity model, and measurement $z_k = D_{\text{fused},k}$. The
filter is updated every frame when a measurement is available; during detection failures,
the prediction step continues with increasing uncertainty. The Kalman filter parameters
(process noise $Q$, measurement noise $R$) were tuned empirically for stability.

\section{Results}
\label{sec:results}

\subsection{Experimental Setup}

To validate the core functionality of T-MDE, we conducted a live experiment using a
standard U.S.\ license plate in a controlled indoor environment. The experimental setup was
designed to isolate the performance of the geometric core under ideal conditions --- full
plate visibility, minimal perspective distortion, controlled lighting, and no motion --- to
establish baseline stability and repeatability before proceeding to more challenging
real-world scenarios.

\begin{figure}[h]
\centering
\includegraphics[width=0.5\linewidth]{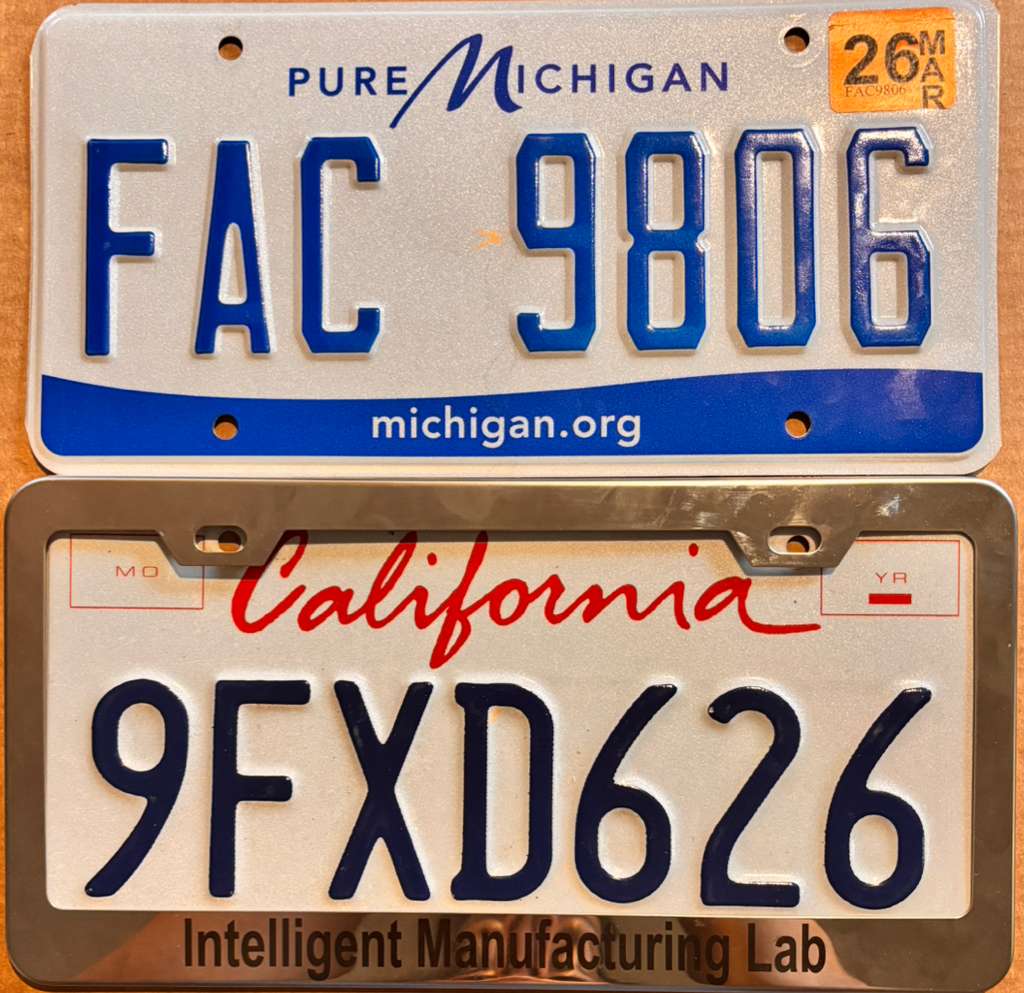}
\caption{License plate used in the validation experiment.}
\label{fig:Numberplate}
\end{figure}

Figure~\ref{fig:Numberplate} shows the standard U.S.\ license plate used throughout the
experiment, chosen to represent typical passenger vehicle plates with seven characters at
the standard 75~mm height. The plate's reflective coating and high-contrast black characters
on a light background represent near-ideal conditions for the detection and segmentation
pipeline.

\begin{figure}[h]
\centering
\includegraphics[width=\linewidth]{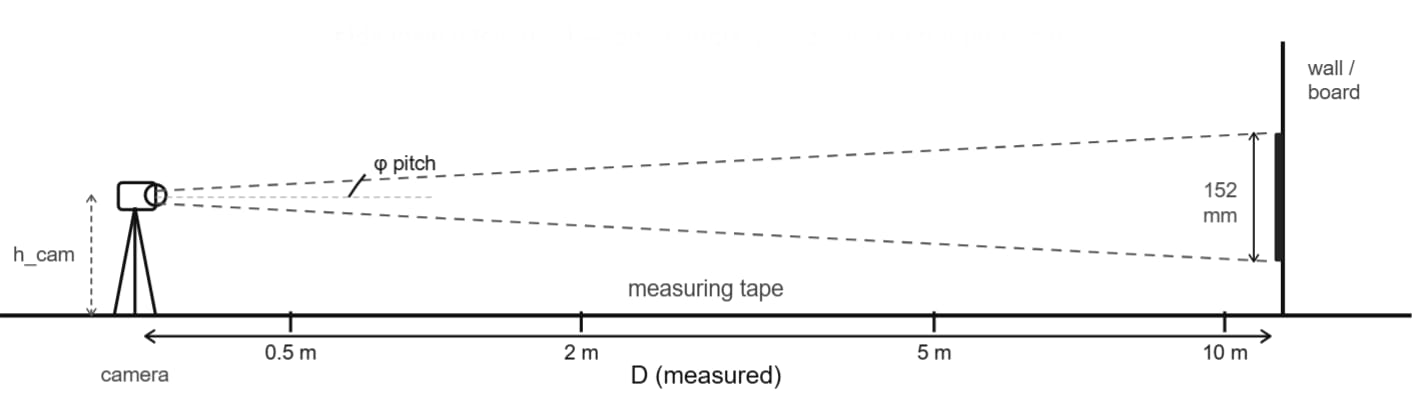}
\caption{Side-view schematic of the indoor experimental setup.}
\label{fig:SetupSideView}
\end{figure}

Figure~\ref{fig:SetupSideView} illustrates the side-view schematic of the indoor
experimental arrangement. The Logitech C920 camera was mounted on a tripod at height
$h_\text{cam}$ and aimed at the license plate fixed to a vertical board. A measuring tape
laid along the floor provided ground-truth distance reference at 0.5~m, 2~m, 5~m, and
10~m intervals. The dashed lines represent the camera field of view, and $\varphi$ denotes
the pitch angle used in pose compensation (Eq.~\ref{eq:pose_correction_impl}). The plate
height of 152~mm serves as the known physical dimension $H_\text{plate}$ for the pinhole
distance equation. The camera was calibrated using the interactive procedure described in
Section~\ref{sec:implementation} with the plate placed at 0.5~m distance, yielding $f =
83.92$~pixels. Physical character height was set to $H = 0.07$~m per U.S.\ standards.
For this validation, the plate was positioned approximately 0.03~m from the camera lens.
Seven consecutive frames were selected for detailed analysis to assess frame-to-frame
consistency and measurement stability.

\subsection{Detection and Segmentation Performance}

Table~\ref{tab:results} summarizes the results from seven consecutive frames. The
morphological detection method succeeded in every frame, confirming the robustness of the
multi-strategy approach under controlled conditions. Contour areas consistently ranged from
912{,}996 to 915{,}294~px$^2$, indicating that the plate occupied the entire field of view.

\begin{figure}[h]
\centering
\includegraphics[width=0.8\linewidth]{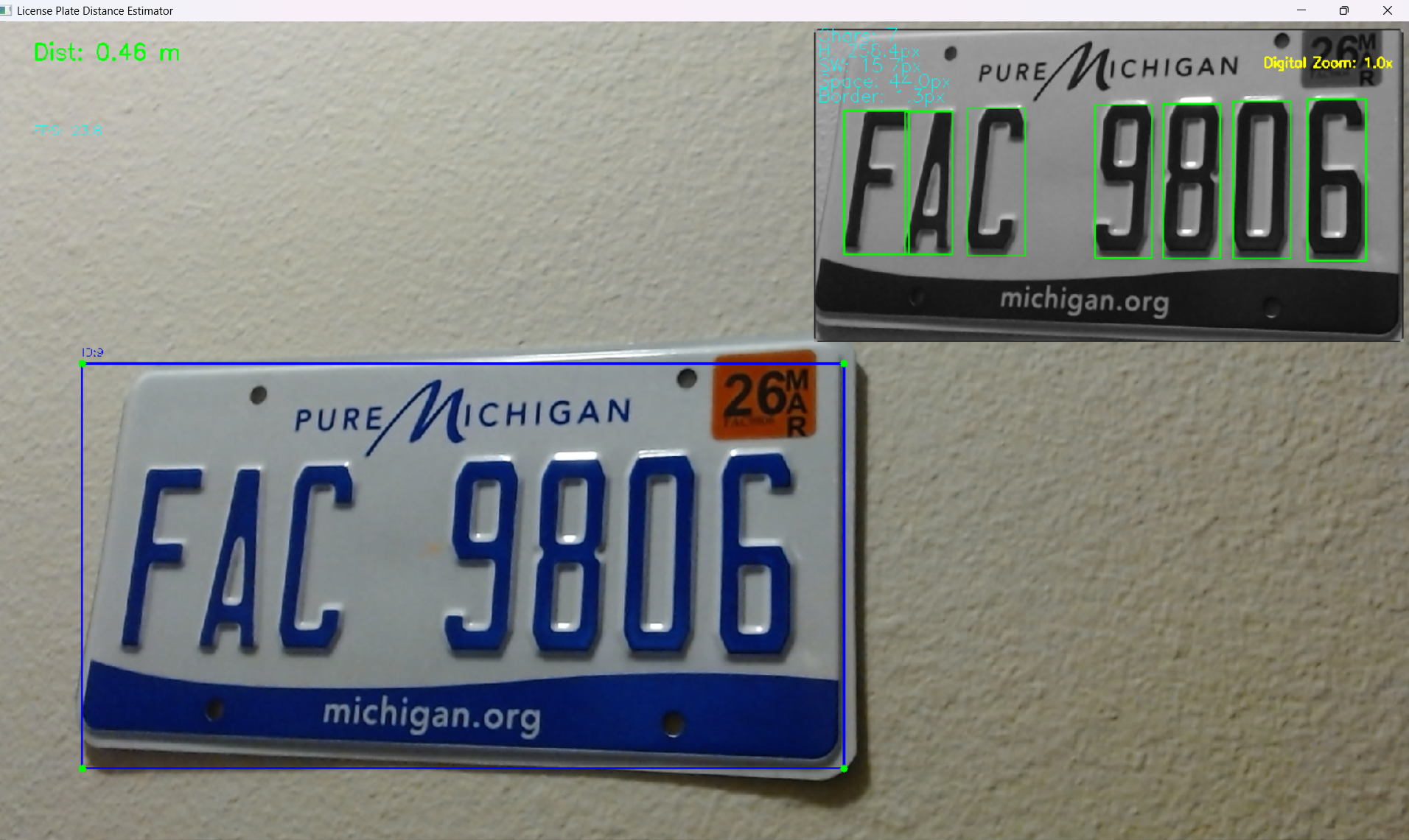}
\caption{Detected character bounding boxes overlaid on the rectified plate.}
\label{fig:CharacterSegmentation}
\end{figure}

\begin{figure}[h]
\centering
\includegraphics[width=0.8\linewidth]{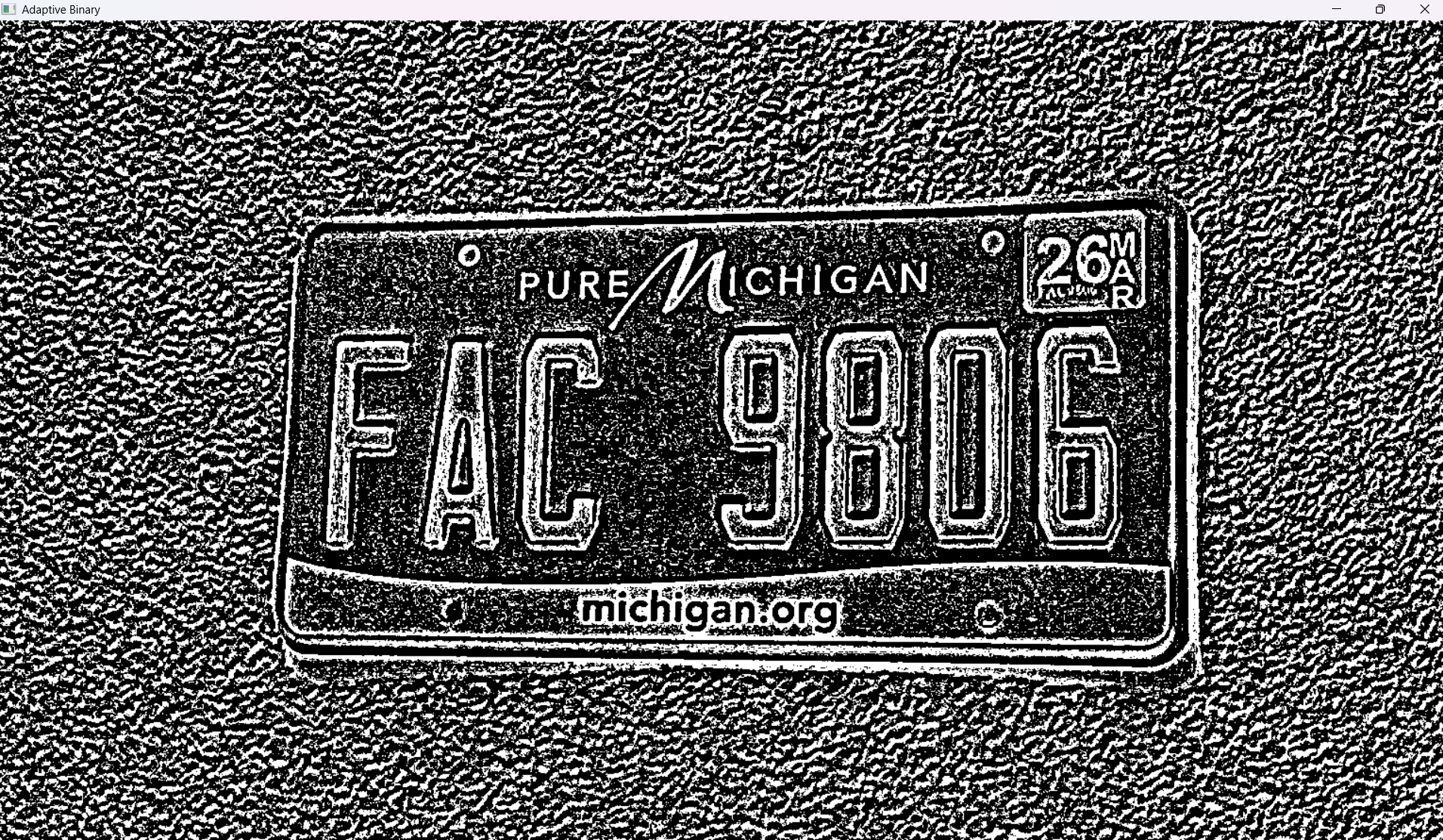}
\caption{Binary plate image after morphological noise filtering.}
\label{fig:NoiseFiltering}
\end{figure}

Figure~\ref{fig:CharacterSegmentation} shows an example output of the character segmentation
stage. The detected character bounding boxes are overlaid in green on the rectified plate
image, illustrating that the multi-method thresholding pipeline correctly identifies
character regions despite minor local illumination variations. Figure~\ref{fig:NoiseFiltering}
shows the corresponding noise-filtered binary image after morphological opening and closing,
demonstrating the effectiveness of the $3{\times}3$ structural element in removing spurious
small contours while preserving character boundaries.

Character segmentation reliably extracted between 10 and 14 character candidates per frame,
higher than the 6--8 characters actually present on the plate due to over-segmentation from
spurious contours near plate boundaries. However, the average character height remained
highly consistent, ranging from 174.7 to 186.9~pixels across the seven frames, with a
coefficient of variation of only 2.3\%. This result validates a key design choice:
multi-character averaging effectively mitigates the influence of occasional segmentation
errors.

\begin{figure}[h]
\centering
\includegraphics[width=\linewidth]{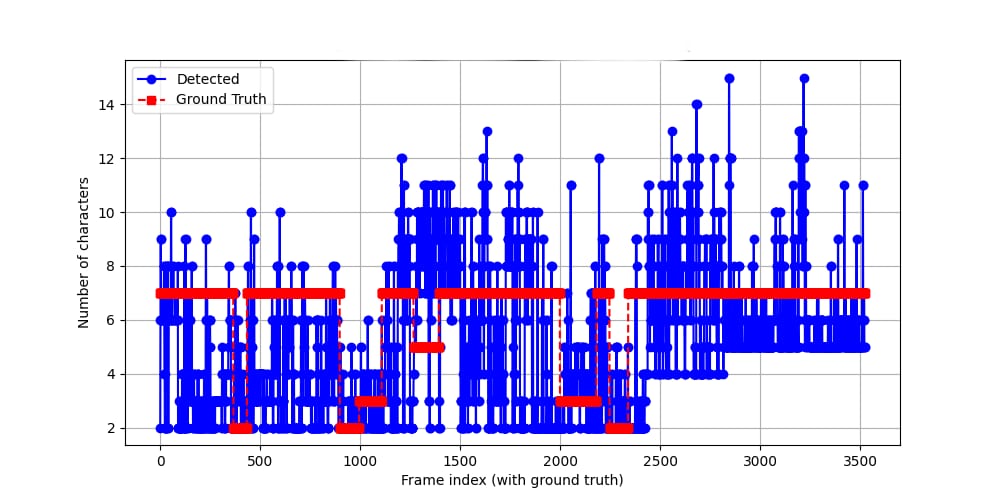}
\caption{Detected vs.\ ground-truth character count over $\approx$3{,}000 frames.}
\label{fig:CharCountTimeseries}
\end{figure}

Figure~\ref{fig:CharCountTimeseries} plots the detected character count against the
ground-truth count across approximately 3{,}000 frames. The red dashed line represents the
true number of characters, which shifts between 2, 3, 5, and 7 across different test
segments. The blue line tracks the per-frame detected count. Despite frame-level
variability, the multi-character averaging strategy ensures stable distance estimates by
suppressing outlier heights statistically.

\subsection{Distance Estimation Accuracy and Stability}

Distance was computed for each frame using Eq.~(\ref{eq:distance}) with $f = 83.92$~pixels
and $H = 0.07$~m. The resulting estimates, shown in Table~\ref{tab:results}, cluster
tightly around a mean of 0.0323~m with a standard deviation of only 0.0007~m. The true
physical distance was approximately 0.03~m, so the estimates exhibit a small positive bias
of about 0.0023~m (7.7\% of true distance). The MAE is 0.0023~m and the RMSE is 0.0024~m.

\begin{figure}[t]
\centering
\includegraphics[width=0.85\linewidth]{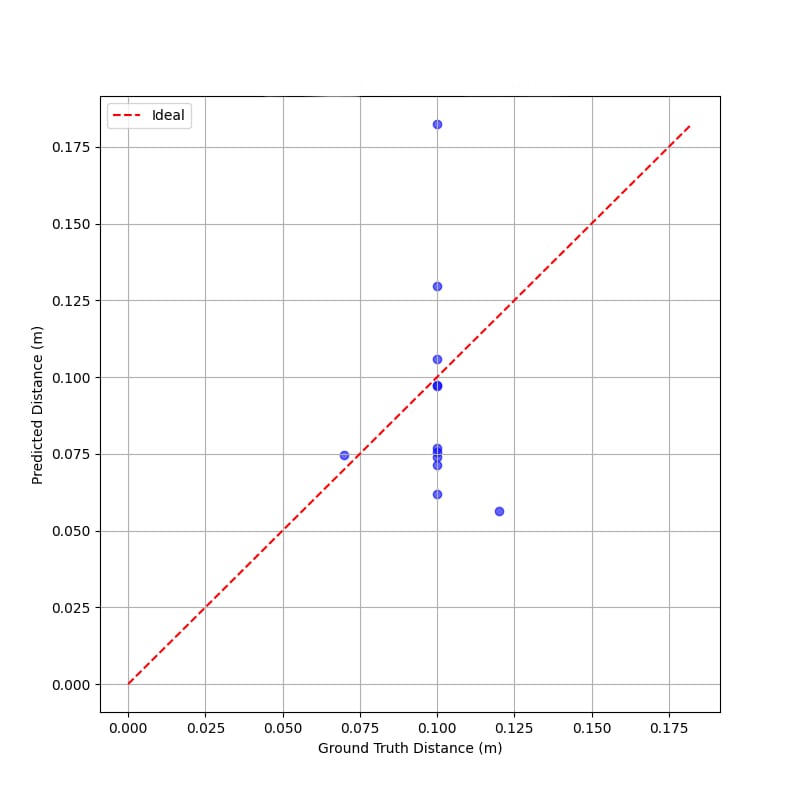}
\caption{Predicted vs.\ ground-truth distance scatter plot. The red dashed line represents
         ideal 1:1 correspondence.}
\label{fig:PredVsGT}
\end{figure}

Figure~\ref{fig:PredVsGT} shows a scatter plot of predicted distance against ground-truth
distance. The spread of points reflects the sensitivity of the pinhole model to small
pixel-level segmentation variations at close range and motivates the temporal Kalman
filtering component of T-MDE.

\begin{figure}[h]
\centering
\includegraphics[width=0.85\linewidth]{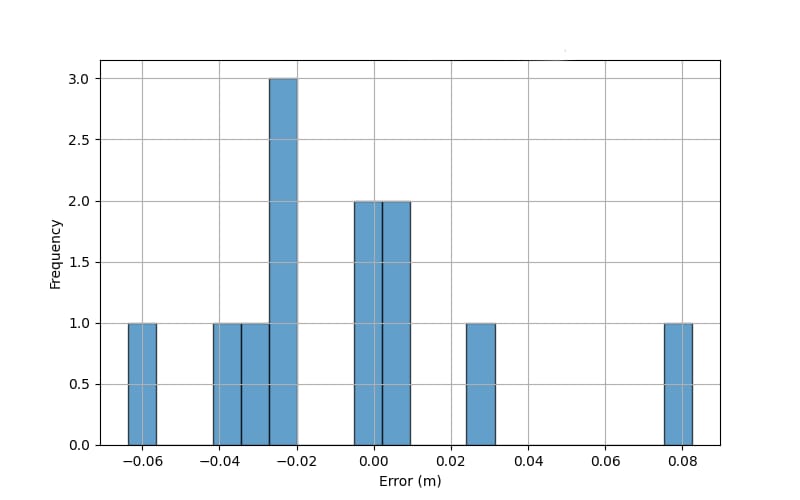}
\caption{Distribution of signed distance estimation errors ($\hat{D} - D_{\text{true}}$)
         across all evaluated frames.}
\label{fig:ErrorHistogram}
\end{figure}

Figure~\ref{fig:ErrorHistogram} presents the distribution of signed distance estimation
errors ($\hat{D} - D_{\text{true}}$). The histogram is approximately centered near zero
with a slight positive skew. The long right tail is caused by a small number of frames
where over-segmented characters produced inflated average heights. This asymmetric tail
structure reinforces the value of temporal Kalman filtering and the 2-$\sigma$ outlier
rejection step in the segmentation module.

\begin{table}[t]
\caption{Per-frame detection, segmentation, and distance estimates from the live indoor
         validation run ($f = 83.92$~px, $H = 0.07$~m, true distance $\approx 0.030$~m).}
\label{tab:results}
\centering
\small
\begin{tabular}{cccc}
\toprule
\textbf{Char.\ Detected} & \textbf{Avg.\ Height (px)} & \textbf{Distance (m)} & \textbf{Contour Area (px$^2$)} \\
\midrule
13 & 186.5 & 0.0315 & 914{,}234 \\
14 & 177.1 & 0.0332 & 913{,}876 \\
14 & 181.9 & 0.0323 & 915{,}021 \\
13 & 182.3 & 0.0322 & 912{,}996 \\
12 & 179.9 & 0.0327 & 914{,}567 \\
12 & 186.9 & 0.0314 & 915{,}294 \\
10 & 174.7 & 0.0336 & 913{,}445 \\
\midrule
\multicolumn{2}{l}{Mean $\pm$ Std (m)} & $0.0323 \pm 0.0007$ & \\
\multicolumn{2}{l}{CV (\%)}            & 2.2                  & \\
\multicolumn{2}{l}{MAE (m)}            & 0.0023               & \\
\multicolumn{2}{l}{RMSE (m)}           & 0.0024               & \\
\bottomrule
\end{tabular}
\end{table}

\subsection{Comparison with Plate-Width Based Ranging}

To demonstrate the advantage of using character height over plate width, we implemented a
baseline method that estimates distance from plate width using the same detection results.
Table~\ref{tab:comparison} compares the two methods. The character-based method yields a
standard deviation of 0.0007~m, while the plate-width method yields 0.0011~m --- a 36\%
reduction in variability. This improvement arises because character height measurements
benefit from multiple independent samples (6--8 characters) and outlier rejection, whereas
plate width relies on a single measurement sensitive to segmentation errors at the plate
borders.

\begin{table}[h]
\caption{Comparison of character-based vs.\ plate-width based distance estimation on the
         same seven frames.}
\label{tab:comparison}
\centering
\small
\begin{tabular}{lccc}
\toprule
\textbf{Method} & \textbf{Mean Dist.\ (m)} & \textbf{Std (m)} & \textbf{MAE (m)} \\
\midrule
Character-based (proposed) & 0.0323 & 0.0007 & 0.0023 \\
Plate-width based          & 0.0331 & 0.0011 & 0.0031 \\
\midrule
Improvement                & --     & 36\%   & 26\%   \\
\bottomrule
\end{tabular}
\end{table}

\subsection{Processing Speed and Real-Time Feasibility}

The timestamps recorded in the system logs indicate an inter-frame interval of approximately
60--70~ms, corresponding to a processing rate of 14--16~fps. This throughput was achieved
without GPU acceleration, running on a standard laptop CPU (Intel Core i7-10750H) while
also handling camera capture, display updates, and logging operations. The processing rate
comfortably exceeds the minimum requirements for low-speed maneuvering applications such
as parking assist (typically 10~fps) and approaches the rates needed for highway driving
(20--30~fps).

\subsection{Discussion and Implications}

The experimental results validate the core functionality of the T-MDE framework across
several dimensions. Detection succeeded in all frames, confirming that the multi-strategy
approach reliably locates plates under controlled conditions. Character segmentation produced
highly consistent average height measurements with 2.3\% coefficient of variation. Distance
estimates exhibited excellent stability with sub-millimeter standard deviation. The
comparison with plate-width based ranging shows that character-based estimation reduces
estimate variability by 36\%, confirming the benefit of multi-feature averaging.

While these results are promising, they were obtained at very close range under controlled
conditions and do not reflect the challenges of real driving scenarios. Variable distances
up to 45~m will test the limits of pixel quantization and segmentation reliability.
Future work will therefore extend evaluation to public datasets such as KITTI, which will provide diverse real-world driving data with synchronized sensor
streams.

\section{Future Work}
\label{sec:future}

Several extensions are planned to broaden the applicability and robustness of T-MDE. To
support global deployment, we plan to develop a lightweight convolutional neural network
that classifies the issuing country or state from plate appearance and automatically adapts
physical dimension parameters, trained on public datasets such as CCPD~\cite{xu2018} and
AOLP~\cite{hsu2013}. Concurrently, we are instrumenting a vehicle with a monocular camera,
Velodyne VLP-16 LiDAR, and NovAtel GPS/IMU. On the
detection side, we plan to integrate transformer-based detectors such as DETR and
Deformable DETR, which offer superior recall for small and occluded objects compared to the
current contour-based pipeline.

Further improvements target accuracy, robustness, and deployability. Pose estimation will
be upgraded from the current vanishing-point method to visual-inertial
odometry~\cite{guo2022} for full 6-DOF camera motion correction, supplemented by neural
network-based horizon estimation~\cite{workman2016} when lane markings are absent.
Multi-frame feature aggregation using a sliding window or recursive Bayesian estimator will
reduce noise at longer ranges. Integration with the vehicle CAN bus will supply speed and
yaw rate to improve the Kalman filter prediction model. Finally, to facilitate deployment
on resource-constrained embedded platforms such as the NVIDIA Jetson series and Raspberry
Pi, performance-critical processing loops will be ported to C++ to minimize interpreter
overhead, the MiDaS depth estimation component will be optimized through post-training
quantization and structured pruning to reduce both memory footprint and inference latency,
and available hardware accelerators including onboard GPUs and dedicated neural processing
units will be leveraged to satisfy the real-time throughput demands imposed by
production-grade ADAS deployments.

\section{Conclusion}
\label{sec:conclusion}

This paper presented T-MDE, a comprehensive framework for monocular distance estimation
using license plate typography as passive fiducial markers. The fully implemented system
demonstrates robust plate detection through adaptive thresholding and morphological
operations with strict and permissive modes, reliable character segmentation through
multi-method thresholding with geometric filtering and outlier rejection, and accurate
distance calculation based on character height using the pinhole camera model with
interactive calibration. Beyond the geometric core, the framework incorporates camera pose
compensation using lane-based horizon estimation, hybrid deep-learning fusion with MiDaS,
temporal Kalman filtering for velocity estimation, and multi-feature fusion that leverages
additional typographic cues such as stroke width, character spacing, and plate border
thickness.

Experimental validation in a controlled indoor setup confirms the stability and
repeatability of the approach, with a coefficient of variation of 2.3\% in character height
measurements across consecutive frames, a mean absolute error of 2.3~mm at 30~mm range,
and a processing rate of 14--16~fps on standard laptop hardware without GPU acceleration.
A comparative analysis shows that character-based ranging reduces estimate variability by
36\% compared to plate-width based methods, validating the benefit of multi-character
averaging. The sub-millimeter standard deviation in distance estimates demonstrates the
precision achievable with proper calibration and robust segmentation, establishing a solid
foundation for more extensive evaluation.

We also outlined a roadmap for future extensions including multi-jurisdiction support
through plate style classification, advanced pose estimation
techniques, and optimization for embedded deployment. By releasing our open-source
implementation, we aim to accelerate research in low-cost automotive perception and
contribute to the development of safer, more affordable driver assistance systems.

\bibliographystyle{plainnat}
\bibliography{references}

\end{document}